# Detection of Thin Boundaries between Different Types of Anomalies in Outlier Detection using Enhanced Neural Networks

Rasoul Kiani[a], Amin Keshavarzi[a,*] and Mahdi Bohlouli[b,c,d]

[a] *Department of Computer Engineering, Marvdasht Branch, Islamic Azad University, Marvdasht, Iran.*

[b] *Department of Computer Science and Information Technology, Institute for advanced Studies in Basic Sciences, Zanjan Iran.*

[c]*Research Center for Basic Sciences and Modern Technologies, Institute for Advanced Studies in Basic Sciences, Zanjan, Iran.*

[d]*Research and Innovation Department, Petanux GmbH, Bonn, Germany.*

[*]Corresponding author: keshavarzi@miau.ac.ir

# Detection of Thin Boundaries between Different Types of Anomalies in Outlier Detection using Enhanced Neural Networks

Outlier detection has received special attention in various fields, mainly for those dealing with machine learning and artificial intelligence. As strong outliers, anomalies are divided into point, contextual and collective outliers. The most important challenges in outlier detection include the thin boundary between the remote points and natural area, the tendency of new data and noise to mimic the real data, unlabelled datasets and different definitions for outliers in different applications. Considering stated challenges, we defined new types of anomalies called Collective Normal Anomaly and Collective Point Anomaly in order to improve a much better detection of the thin boundary between different types of anomalies. Basic domain-independent methods are introduced to detect these defined anomalies in both unsupervised and supervised datasets. The Multi-Layer Perceptron Neural Network is enhanced using the Genetic Algorithm to detect new defined anomalies with a higher precision so as to ensure a test error less than that calculated for the conventional Multi-Layer Perceptron Neural Network. Experimental results on benchmark datasets indicated reduced error of anomaly detection process in comparison to baselines.

## Introduction

Data extraction (Keshavarzi et al., 2008) and pre-processing operations lead to a refined explorable dataset in different machine learning applications such as cloud computing (Keshavarzi et al, 2019; Keshavarzi et al., 2017), big data (Bohlouli et al., 2013), and sensor networks (Jafarizadeh et al., 2017). Pre-processing aims at identification and removing outliers to improve the quality of cleansing process (Agarwal, 2013; Kiani et al.,2015). Outliers show a higher deviation and are not in line with the behaviour of general dataset, which could cause unexpected results in analytics. Outliers probably created due to measurement error, the inherent variability of data or faulty sensors (Chandola et al., 2009; Agarwal, 2015; Chandarana and Dhamecha, 2015). Noises are weak outliers but anomalies are strong outliers. The boundary between the noises and anomalies is not clear but can be determined through different analytical methods (Agarwal, 2015). Anomalies are divided into three categories of point, contextual and collective anomalies (Chandola et al., 2009; Agarwal, 2015; Song et al., 2007; Malik et al., 2014).

Point Anomalies (PA) are located at a considerable distance from normal data and diverged from the usual pattern of data. According to conditions, contextual anomalies can be (or not to be) outlier relative to normal data. Collective anomalies are a set of related outliers relative to normal data. Such anomalies may be free of deviations alone (Chandola et al., 2009; Malik et al., 2014; Gupta et al., 2014 ). In fact, point and collective anomalies are two subsets of contextual anomalies (Chandola et al., 2009; Agarwal, 2015). Based on the use of labelled data, outlier detection approaches are divided into supervised, semi-supervised and unsupervised methods (Vijayarani and Nithya, 2011; Theiler and Cai, 2003; Steinwart et al., 2005; Fujimaki et al., 2005; Bolton et al., 2001).

Also outlier detection methods are divided into distribution-based, clustering-based, distance-based and density-based methods (Hodge and Austin, 2004; Kou et al., 2007; Zhang, 2008; Chllalagalla et al., 2010). The key components of anomaly detection methods are research area, anomaly detection technique, problem characteristic and application domain (Chandola, 2009).

The most important challenges in outlier detection include the thin boundary between the remote points and natural area, the tendency of new data and noise to mimic the real data, unlabelled datasets and different definitions for outliers in different application areas (Chandola, 2009; Agarwal, 2015). In this paper, the thin boundary between normal data and various types of anomalies is examined. Furthermore, other types of anomalies called Collective Normal Anomaly (CNA) and Collective Point Anomaly (CPA) are investigated.

- CNA: There is a thin boundary between Normal Data (ND) and CNA. Due to the characteristics of ND, it is assumed that CNA can be clustered. CNA is a cluster that its standard deviation density is greater than or equal to the threshold for standard deviation of all clusters.
- CPA: CPA is a subset of Point Anomaly (PA) and there is a thin boundary between PA and CPA. Due to the characteristics of PA, it is assumed that CPA cannot be clustered.

Figure 1 shows ND, CNA, PA and CPA in a schematic plot and the thin boundary between various types of anomalies is visible.

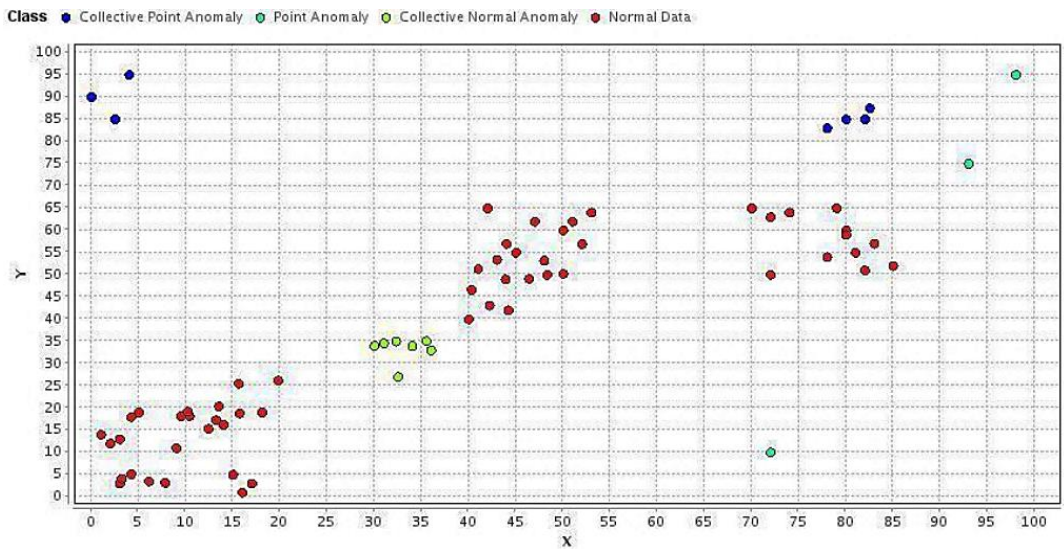

Figure 1. A schematic plot of thin boundary between normal data and various types of anomalies.

For this purpose, unsupervised and supervised datasets are first studied. Using proposed framework in this paper, the supervised dataset is divided into subsets based on the number of classes. The Multi-Layer Perceptron Neural Network (MLP-NN) is also improved using the Genetic Algorithm (GA) to detect the thin boundary between different types of anomalies. Because of the fact that neural network learning which is based on neurons weight and detection accuracy is variable in each epoch, using GA seems possible to solve this problem which is improved both better detection of new defined anomalies and reducing the test error.

The rest of this paper is organized as fallows. Section 2 reviews related work in outlier detection. The proposed method is discussed in detail in Section 3. The results are analyzed in Section 4. Finally, conclusions are presented.

**Literature Review**

As a supervised or semi-supervised method, the neural networks have been used to detect outliers and anomalies in various fields such as host based intrusion detection (Ghosh et al., 1998), network intrusion detection (Ramadas et al., 2003; Smith et al., 2002; Zhang et al., 2001), credit card fraud detection (Zhang et al., 2001; Aleskerov et al., 1997), mobile fraud detection (Barson et al.,1996; Taniguchi et al., 1998), medical and public health domain (Campbell, 2001), fault detection in mechanical units (Diaz and Hollmen, 2002; Li et al., 2002), structural damage detection (Shon et al., 2001), image processing (Singh et al., 2004; Augusteijn and Folkert, 2002) and anomalous topic detection in text data (Manevitz and Yousef, 2001). Figure 2 shows the variation of efficiency with dimensions for all methods. Moreover, Figure 3 shows the variation of scalability with the dimensions respectively for all methods (Malik et al., 2014). As can be seen in Figures 2 and 3, both efficiency and scalability are considered based on dimensionality respectively for all the methods. It seems likely the methods based on NN and clustering benefit greatly from the best efficiency and scalability. For example, when dimensionality is 80, the scalability of NN, clustering and density methods are approximately equal therefore NN seems a much better choice owing to the fact that its efficiency is better.

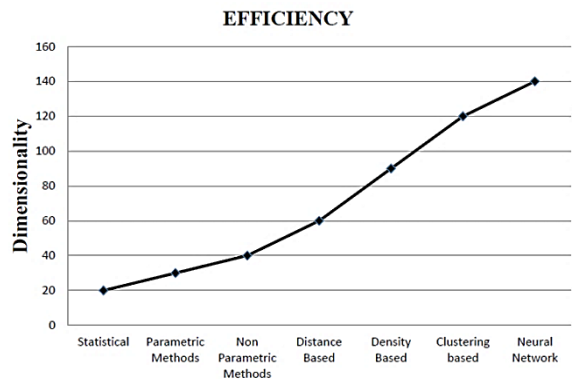

Figure 2. Efficiency of various outlier detection methods in scale up (Malik et al., 2014).

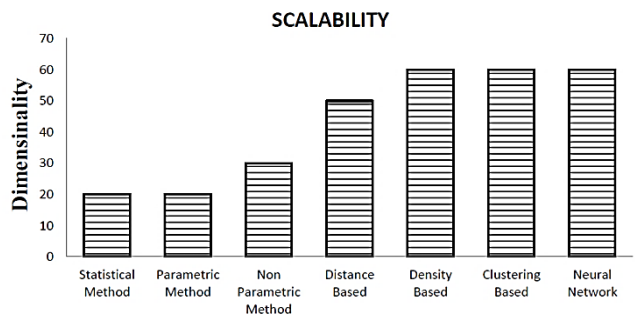

Figure 3. . Scalability of various outlier detection methods in scale up (Malik et al., 2014).

A 3-step process has been proposed to detect false alarms and outliers (Hachmi et al., 2015). In the first step, preliminary alerts are clustered to create a set of meta-alerts. In the second step, outliers are removed from the meta-alerts. In the third step, a binary classification algorithm is used to classify meta-alerts into attacks and false alarms. An extended statistical unsupervised method has been used to detect outliers in object-relational data (Riahi et al., 2015). For this purpose, a metric was introduced based on the likelihood ratio of vectors of population association and individual association. To detect outliers in large matrices, a two-stage adaptive approach has been suggested that its performance is guaranteed using an inference met (Li et al., 2015). Song et al., (2007) proposed a general-purpose method called conditional anomaly detection. They used three different learning algorithms for their proposed model. A distributed outlier detector with a reasonable speed and efficiency has been proposed to detect so-called global outliers in a distributed database (Zhang et al., 2012). Wang and Davidson (2009) detected contextual outliers by using the random walks graph. The most important feature of this approach is to consider scores for outliers. A data driven approach has been proposed to detect anomalies in the patient management actions (Hauskrecht et al., 2010). This method is based on past patient records in the electronic patient health record system. A semi-supervised framework based on fixed-background mixture has been proposed to detect anomalies (Vatanen et al., 2012). This framework is robust enough to detect patterns of anomaly model. To detect collective anomalies and DoS attacks in network traffic analysis, a framework has been suggested based on X-means clustering algorithm (Ahmed and Mahmood, 2014). Noble and Cook (2003) used anomalous infrastructure detection and anomalous sub graph detection to provide a graph-based approach for anomaly detection. Yang and Liu (2011) detected anomalies in collective moving patterns using the hidden Markov model. Abnormal detection research preprocessing the data and sets the normal sample set has been presented. This method based on outlier mining calculated the outlier score of each sample in the normal sample set (Zhang et al., 2018). Taylor et al. proposed the outlier detection method using the super efficiency which removal of one outlier little effect in estimation since the neighbor outlier serves as a proxy benchmark. In other words, they developed an alternative method based on the stochastic DEA model of Banker (Boyd et al., 2016). Ko et al., (2017) suggested the model based on data integration and machine learning-based anomaly detection so as to the overcome the conventional methods for estimating the level of quality. Also, the method for segmentation and indexing multi-dimensional time series data is introduced. Maheshwari and Singh (2016) proposed an algorithm to output clusters and outliers in a divide and conquer manner. The method following outliers in each cluster identified core objects outliers. Guo et al. (2018) proposed a new distance-based method on which depends the data structure to detects such points. In the proposed method, firstly, a global binary tree is used and then the local distance score of point is calculated for evaluating to what degree the observations in an outlier. Zhao and Hryniewicki (2018) proposed an algorithm called XGBOD which was a new semi-supervised method. XCBOD described and demonstrated for enhanced detection of outliers from normal data. This framework combined the strengths of both supervised and unsupervised methods by a hybrid approach. Kutsuna and Yamamoto (2017) suggested a novel method for outlier detection using binary decision diagram which is used a new measure for detecting outliers. Lin et al., (2018) proposed a method has employed a spatial-feature-temporal tensor model analyzed latent mobility patterns through unsupervised learning and LOF algorithm is used to localize anomaly in a given time interval. Macha and Akoglu (2018) proposed a new approach called x-PACS which are used reverse engineering to detect anomalies based on both the groups and characterizing subcase and features rules.

## The Proposed Method

*Assumptions*

It is assumed that there are two types of datasets: (1) datasets containing labeled data and (2) those containing unlabeled data. The main assumption of our proposed method is that data has been previously labeled using common techniques such as clustering algorithm, decision tree, hidden Markov Model, etc. Equation (1) shows the relationship between different types of anomalies and normal data (Chandola et al., 2009):

- PA: If an individual data instance can be considered as anomalous with respect to the rest of data, then the instance is termed as a PA.
- Collective Anomalies (CA): If a collection of related data instances is anomalous with respect to the entire data set, it is termed as a CA. The individual data instances in a collective anomaly may not be anomalies by themselves, but their occurrence together as a collection is anomalous.
- ND: ND instances occur in dense neighborhoods, while anomalies occur far from their closest neighbors.

The relationship between the PA and CPA is shown in Equation (2). Based on Equation (2) CPA is a subset of PA and there is a thin boundary between PA and CPA. Equation (3) defines CPA that the neighborhood radius of CPA is less than average neighborhood radius of PA.

$$PA \cup CA \cup ND = Dataset \qquad (1)$$

$$CPA \subset PA \qquad (2)$$

$$CPA = \{P_i \in CPA | P_i \in PA \text{ and } Out\_Rad_{P_i} < Out\_Rad_{PA}\} \qquad (3)$$

where $Out\_Rad_{Pi}$ is the neighborhood radius of $P_i$ as well as its average distance to PA and $Out\_Rad_{PA}$ is the neighborhood radius of PA.

Equations (4)-(6) show the calculation of neighborhood radius.

$$Out\_Rad_{Pi} = \frac{\sum_{i=1}^{k} MDist_{O_i}}{k} \qquad (4)$$

$$MDist_{O_i} = \left\{ \frac{\sum_{j=1}^{k} Dist(O_i, O_j)}{k-1} \middle| i \neq j \;,\; i=\{1,...,k\} \right\} \qquad (5)$$

$$Dist(O_i, O_j) = \left\{ \sqrt{(x_{O_i} - x_{O_j})^2 + (y_{O_i} - y_{O_j})^2} \middle/ \; i,j=\{1,...,k\} \right\} \qquad (6)$$

where $Out\_Rad_{Pi}$ is the outlier radius of CPA, $MDist$ is the mean distances table from point anomalies, $O_i$ and $O_j$ are PA, and $k$ is the number of point anomalies.

Equation (7) shows the relationship between the ND and CNA. Based on Equation (7) CNA is a subset of ND and there is a thin boundary between ND and CNA. Due to the characteristics of ND means that their neighborhood radius is less than the mean distances from points, is assumed that CNA can be clustered in order to use in supervised data. The definition of CNA is given in Equation (8). CNA is a cluster that its standard deviation density is greater than or equal to the threshold for standard deviation of all clusters.

$$CNA \subset ND \tag{7}$$

$$CNA = \{C_i \in C | C_i \in ND \text{ and } \sigma\_Den_{C_i} \geq Th\_\sigma_{Den_C}\} \tag{8}$$

where $C_i$ is one of detected clusters, $C$ is the set of all clusters, $\sigma\_Den_{C_i}$ is standard deviation of cluster density $C_i$, and $Th\_\sigma_{Den_C}$ is the threshold for standard deviation of all clusters.

In this paper the research area is data mining as well as this the application range is independent of domain and the problem characteristic is the type of anomaly .The proposed method to detect different types of anomalies is described and a new framework is proposed for labeling supervised datasets in the proposed framework section. After that, MLP-NN is enhanced using the GA to increase the precision of anomaly detection.

## *The Proposed Framework*

One of the important points considered in this paper is adaptability of the proposed algorithm with both supervised and unsupervised datasets. As previously mentioned anomalies have been labeled using common techniques and are ready to be used in the neural network. In the first step, the supervised dataset is divided into sub datasets to detect local anomalies.

*Step 1:* Various types of anomalies should be investigated in all classes in the supervised dataset. Thus, among $k$ features in the reference dataset, $l$ features with a higher separation capability should be analytically selected as the main features. In other words, based on three criteria including the type of dataset, functional domain and its features, the most distinguishing features should be selected. Although $l$ features have been selected to suit all classes means that supervised dataset, they may be not the best if they are evaluated locally (in each sub dataset). Thus, aggregation technique is used here. That is to say, aggregation is a type of data smoothing. Therefore, two aggregation techniques are used to reduce the number of features from $k$ to $l$. As a first technique, normalization is used to improve the accuracy of data mining algorithms. The second technique is to weight for valuation of all features. The range of numbers has a direct effect on the weight obtained for each feature in the weighting process. If normalization techniques are not used, weights will be unbalanced leading to unsmoothed features. Accordingly, normalization technique is used to solve this problem to put all the numbers for all features in a constant range.

Equation (9) shows the weighting formula and Equation (10) shows the formula for constructing new features.

$$W_i = \frac{X_i^1 + X_i^2 + \ldots + X_i^j}{j} \quad i = 1\ldots n, j = 1\ldots m \text{ and } m = k - l \tag{9}$$

where $W_i$ is the weight of sample $i$, $i$ is the sample number, $n$ is the total number of samples in the dataset, $j$ is the feature number that its weight is calculated, $k$ is the total number of features in the dataset. It should be noted that $m=k-l$ since when $l$ main features are considered as a main features globally, they do not have the best features locally in each sub datasets, and $X$ is the normalized sample.

$$Att\_New_i^j = W_i + Att_i^j \qquad i=1\ldots n, j=1\ldots l \qquad (10)$$

where $Att\_New$ is a new feature for each sample in the dataset, $i$ is the number of sample and $j$ is the number of feature.

*Step 2:* Dividing datasets into sub datasets based on the number of classes in supervised dataset, and apply clustering algorithm for unsupervised dataset.

*Step 3:* Labeling various types of anomalies.

*Step 4:* Normalization: since data in sub datasets are affected by the weights used in the Equation (9), normalization is applied at this stage.

*Step 5:* Integration of sub datasets.

Figure 4 shows the proposed framework for labeling the supervised dataset.

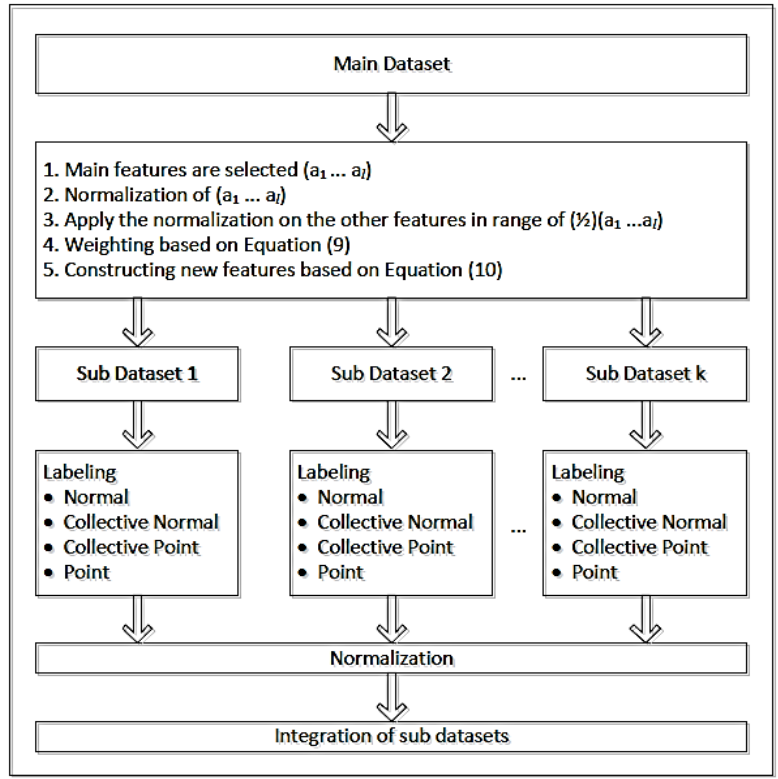

Figure 4. The proposed framework for labeling the supervised dataset.

## MLP neural network Optimization using Genetic Algorithm

MLP-NN and back propagation learning algorithms are used to detect anomalies in supervised and unsupervised datasets. As a drawback, this type of neural network gives different precision, recall and detection error values each time it is applied to detect various types of anomalies with same inputs. By enhancing this type of neural network, its detection performance will be improved compared to MLP-NN with conventional back propagation learning algorithm.

Two MLP-NN have been used in the overall scheme, one separately and the other as the fitness function in GA, but all the initial parameters are the same. Table 1 lists the MLP-NN parameters.

**Table 1.** MLP neural network parameters.

| Row | Parameter | Value |
|---|---|---|
| 1 | Input Weight & Bias | Generated by GA |
| 2 | Weight & Bias Range | [0-1] |
| 3 | Input Neuron | 2 |
| 4 | Output Neuron | 4 |
| 5 | Layer Number | 2 (1 Hidden Layer & 1 output Layer) |
| 6 | Hidden Layer Function | Tansig |
| 7 | Hidden Layer Size | 10 |
| 8 | Output Layer Function | Tansig |
| 9 | Initial Function | Initlay |
| 10 | Perform Function | Mse |
| 11 | Train Function | Trainscg |
| 12 | Learning Algorithm | Back Propagation |

Different parameters used in the NN are selected according to the application range and the optimal adaptability.

1. The goal is to improve the results of the neural network, so both networks have the same initial input weights and this is why this is done using the genetic algorithm.
2. Selecting an appropriate matrix for weights and biases leads to rapid convergence of the neural network but improper selection leads to local optima. This is why they are considered in the range of 0 to 1.
3. Since the data are shown in a 2-dimensional space, 2 input neurons are considered, and the number of neurons increased.
4. Since the data are divided into 4 groups, 4 neurons are considered in the output layer so that the output of each neuron can be 0 or 1. To show anomalies and ND, one of the output neurons is 1 and the other neurons are 0.
5. Although an increase in the number of hidden layers increases the learning ability, calculations in the training and testing steps will increase. Most problems (models) that cannot be separated linearly (using a line in a 2-dimensional space), can be solved by 2 to 3 layers in the MLP network (1 output layer, one or two hidden layers). This is why the number of network layers is 2.
6. In the MLP-NN, neurons activation function in the hidden layers must be of Sigmoid type. Otherwise, the MLP-NN becomes a single-layer perceptron neural network and cannot detect non-linear inseparable problems. There are two types of Sigmoid function including:
    a. Tansig: Hyperbolic tangent sigmoid transfer function. Tansig is a neural transfer function. Transfer functions calculate a layer's output from its net input. This is mathematically equivalent to tanh(N). It differs in that it runs

faster than the MATLAB implementation of tanh, but the results can have very small numerical differences. This function is a good tradeoff for neural networks, where speed important and the exact shape of the transfer function is not.

    b. Logsig: Log-sigmoid transfer function. Logsig is a transfer function.

Therefore, Tansig is used.

7. The small number of neurons in the hidden layer causes inadaptability while the large number of neurons in the hidden layer leads to over fitting. Therefore, 10 neurons were considered in the hidden layer by trial and error.
8. The use of Sigmoid function in the output layer limits the network output to a small range. As previously stated, Tansig function is more appropriate for this purpose.
9. Since the weights and biases are injected to the neural networks, Initlay (Layer-by-layer network initialization) function is used. Initlay is a network initialization function that initializes each layer $i$ according to its own initialization function net and returns the network with each layer updated. Initlay does not have any initialization parameters.
10. MSE (Mean squared normalized error performance function) is used as the performance function. MSE is a network performance function which measures the network's performance according to the mean of squared errors and returns the mean squared error. Note that MSE can be called with only one argument because the other arguments are ignored. MSE supports those ignored arguments to conform to the standard performance function argument list.
11. To select training algorithm for the MLP-NN, different parameters such as problem complexity, the number of data in the dataset, the number of weights and biases, the error and so on should be considered. According to the parameters listed above, Trainscg (Scaled conjugate gradient back propagation) function is used which is a network training function that updates weight and bias values according to the scaled conjugate gradient method. Trainscg can train any network as long as its weight, net input, and transfer functions have derivative functions. Back propagation is used to calculate derivatives of performance with respect to the weight and bias variables. One of the main reasons for selecting trainscg function is to improve network generalization as well as this one of the ways to improve the network generalization is early stopping where the dataset is divided into training, evaluation and testing data and the trainscg function shows a better performance with early stopping.

This section outlines the GA steps to enhance the results of MLP-NN.

*Step 1:* the initial population is generated by the GA. The number of genes in individuals equals the number of weights and biases required for the MLP-NN. The purpose is to apply the same input weights to the MLP-NN outside the GA and the evaluation function (the MLP-NN inside the GA). The extracted weights are generated as an initial population in the form of a matrix where the number of rows equals the population in each generation and the number of columns is equal to the total number of weights and biases. In the future generations, GA will produce the next generation.

*Step 2:* Applying the crossover operator according to Equation (11).

$$\text{Infant}_i = \begin{cases} \{G_1^i, \ldots, G_k^i\} = \{X_1^i, \ldots, X_k^i\} \text{ where} \\ \quad i \in \{1, 2\} \\ \{G_k^i, \ldots, G_n^i\} = (X_j^i \times a) + (X_j^{i\pm 1} \times (1-a)) \text{ where} \\ \quad j \in \{k, \ldots, n\} \\ \quad k \in \{2, \ldots, n-1\} \\ X_j^{i\pm 1} = X_j^{i+1} \mid i=1 \;,\; X_j^{i\pm 1} = X_j^{i-1} \mid i=2 \end{cases} \quad (11)$$

where $\text{Infant}_i$ is the $i^{th}$ infant, $Gj^i$ is $j^{th}$ gene of the $i^{th}$ infant, $k$ is a random number, $X_j^i$ is $j^{th}$ gene of the $i^{th}$ parent, $n$ the total number of genes equal to the number of weights and biases of the MLP-NN.

*Step 3:* The mutation operator (Equation (12)) is used to search in a larger space to avoid local optima.

$$\text{Infant}_i^{New} = \begin{cases} \text{Infant}_i^{Old} \text{ where} \\ \quad Prob_1 < Mut.Rate \\ \{X_1^{Old}, \ldots, X_j^{New}, \ldots, X_n^{Old}\} \text{ where} \\ Prob_1 \geq Mut.Rate \;,\; Prob_2 = j \in \{1, \ldots, n\} \\ X_j^{New} = X_j^{Old} + (k \times Prob_3) \text{ where} \\ 0 < Prob_3 < 1, \begin{cases} k = -1 \mid Prob_4 < 0.5 \\ k = +1 \mid Prob_4 \geq 0.5 \end{cases} \end{cases} \quad (12)$$

where $Prob_1$, $Prob_2$, $Prob_3$ are probabilistic values, $\text{Infant}_i^{New}$ is the $i^{th}$ mutated infant, $\text{Infant}_i^{Old}$ the current un-mutated infant, $Prob_i$ mutation probability of the $i^{th}$ infant, $Mut.Rate$ desired mutation rate, $X_j^{Old}$ $j^{th}$ un-mutated gene of the $i^{th}$ infant, $X_j^{New}$ is $j^{th}$ mutated gene of the $i^{th}$ infant, $Prob_2$ is the number of gene to mutate the $i^{th}$ infant, $Prob_3$ the change in the mutated gene and $Prob_4$ is increased or decreased change in the mutated gene.

*Step 4:* The evaluation function is used to make decisions for the next generation. For this purpose, the MLP-NN with the same parameters of the MLP-NN outside the genetic algorithm is used. The fitness function is defined as follows:

$$Fitness = Final\ Test\ Error = Average\ of\ Target\ Test\ Error \quad (13)$$

*Step 5:* The selection function is applied. Assuming a selection rate of 70%, 70% of the best errors are selected and the rest are selected randomly from the normal data. A ranking-based selection procedure is used, because a member with low adaptability may have appropriate and effective genes.

*Step 6:* One of the following conditions will end the algorithm.
- The implementation cycles of the GA.
- Reaching the minimal error shown by Goal.

*Step 7:* At the end, the test errors obtained from the enhanced and conventional MLP-NN are compared.
Table 2 lists the initial parameters of the GA.

**Table 2.** The initial parameters of the genetic algorithm.

| Row | Parameter | Value |
|---|---|---|
| 1 | Cycle | **20** |
| 2 | Population Size | **15** |
| 3 | Crossover α | **0.3** |
| 4 | Mutation Rate | **0.1** |
| 5 | Selection Rate | **0.7** |
| 6 | Goal | **0** |
| 7 | Fitness Function | **Test Error** |

The reasons for selecting the initial parameters of the GA are discussed.
1. The cycle is selected by trial and error.
2. Initial population is selected by trial and error.
3. The use of crossover operator generates members with adaptability higher than the average and this avoids dispersion. For this purpose, a single point is used. An increase in the number of points in the crossover operator will result in higher variation in the search space and a lower reliability (the answers will considerably change in different generations).
4. Mutation leads to search in the space that has not been previously investigated. Mutation rate should not be high, because the GA becomes a completely random search algorithm and thus convergence is delayed.
5. One of the problems with small population in the GA is local optima. To overcome this problem, Rank Scaling selection function is used. The default fitness scaling option, Rank, scales the raw scores based on the rank of each individual instead of its score. The rank of an individual is its position in the sorted scores: the rank of the fit individual is 1, the next most fit is 2, and so on. The rank scaling function assigns scaled values so that. Rank fitness scaling removes the effect of the spread of the raw scores.
6. The target error of the fitting function is 0. When an error of 0 is achieved, the algorithm is stopped.
7. The fitness function in the GA is defined using MLP-NN to minimize the test error. Figure 5 shows the proposed scheme to enhance MLP-NN using the GA. It should be noted that the MLP-NN is enhanced using the GA to detect new defined anomalies with a higher precision so as to ensure a test error less than that calculated for the conventional MLP-NN.

**Results and Discussion**

The proposed techniques have applied on 2 parts (Part A and B). Firstly, three datasets were selected based on an idea which is showed the evaluation parameters include precession, recall, test error and ROC curve. Secondly, the ability of proposed framework so as to detect the thin boundary challenge between new anomalies based on 8 UCI datasets has considered. Additionally, we have used a few benchmark datasets based on the repository which proposed in (Campos et al., 2016) to calculate both true positive rate and false positive rate. It should be noted that datasets have picked in various fields.

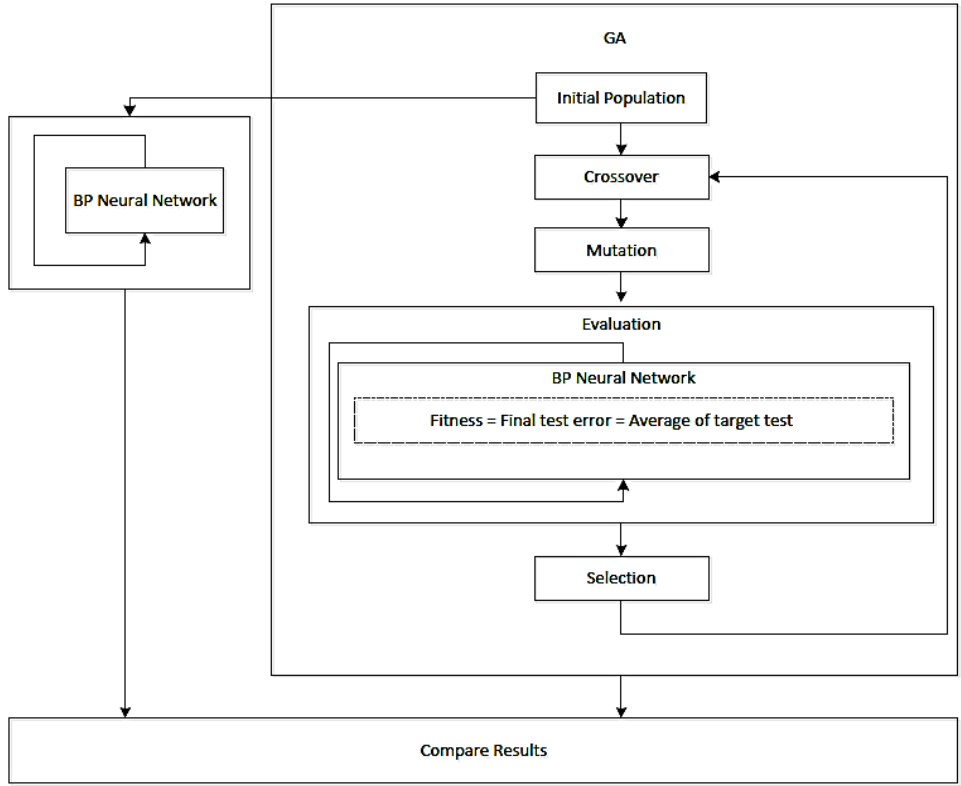

**Figure 5.** The proposed scheme to enhance MLP- NN using the GA.

## Part A

The proposed techniques have applied on three datasets. The first dataset has been randomly generated and its data are labeled according to Table 3 after applying a clustering algorithm. Figure 6 shows the distribution of the first dataset in a 2-dimensional space. Using the confusion matrix, the results of the enhanced and conventional MLP-NN are compared.

**Table 3.** Labeling data in the first dataset.

|  |  |  |  |  | 195 | #Point |
|---|---|---|---|---|---|---|
| Cluster 5 | Cluster 4 | Cluster 3 | Cluster 2 | Cluster 1 | 5 | #Cluster |
| #CNA 3 | #CNA 12 | #CNA 12 | #ND 25 | #ND 48 |  |  |
|  |  |  |  |  | 73 | #ND |
|  |  |  |  |  | 27 | #CNA |
|  |  |  |  |  | 53 | #CPA |
|  |  |  |  |  | 42 | #PA |

According to the matrix (Figures 7 and 8), the test error of the MLP-NN enhanced by the GA is 10% while the corresponding error for the conventional MLP-NN is 26.7%. Here, 1 represents the ND, 2 the CNA, 3 the CPA and 4 represents PA. The most important thing is high-precision detection of the thin boundary between various types of anomalies as is visible in the confusion matrix.

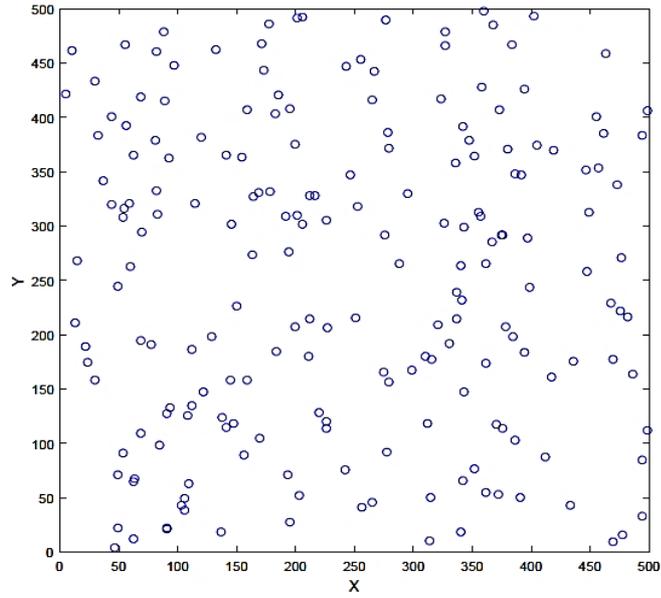

**Figure 6.** Distribution of the first dataset in a 2-dimensional space.

**Figure 7.** Neural network confusion matrix for the first dataset.

**Figure 8.** Genetic algorithm confusion matrix for the first dataset.

The second dataset is used to apply the proposed techniques in (Rehm et al., 2007). Table 4 shows the labelled data in the second dataset. Figure 9 shows distribution of data in the second dataset in a 2-dimensional space.

**Table 4.** Labeling data in the second dataset.

|  | Cluster 3 | Cluster 2 | Cluster 1 | 45 | **#point** |
|---|---|---|---|---|---|
|  | #CNA 4 | #ND 9 | #ND 13 | 3 | **#Cluster** |
|  |  |  |  | 22 | **#ND** |
|  |  |  |  | 4 | **#CNA** |
|  |  |  |  | 13 | **#CPA** |
|  |  |  |  | 6 | **#PA** |

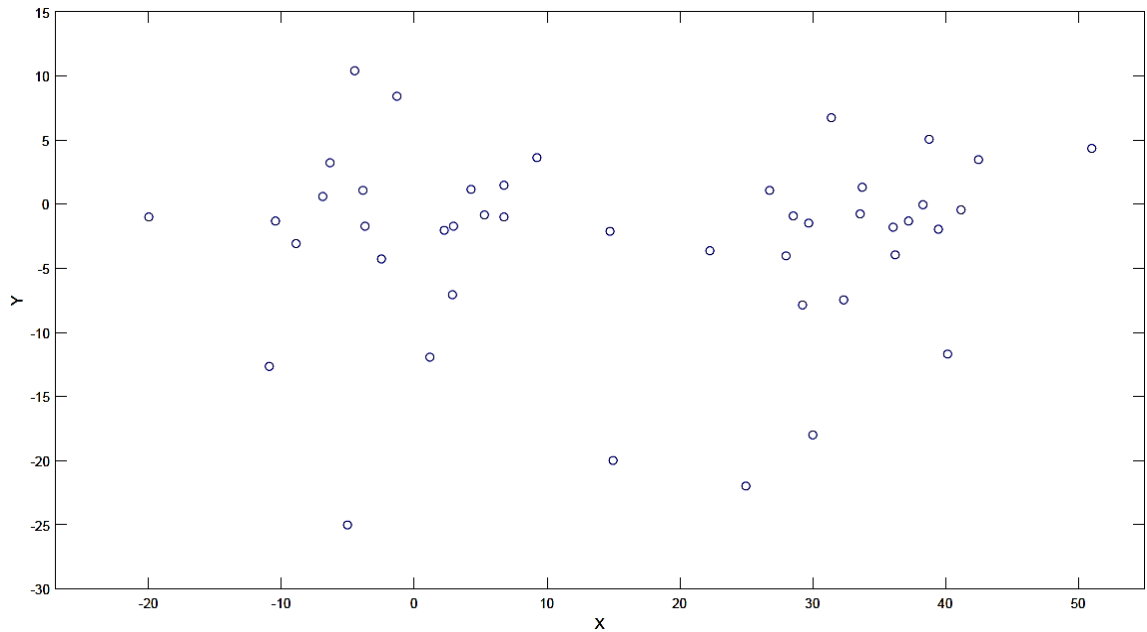

Figure 9. Distribution of data in the second dataset in a 2-dimensional space.

Both Figures 10 and 11 show the confusion matrix for the second dataset. The reasons why no data is not selected for the Class 2 in the testing step are obvious:(1) insufficient data, (2) the procedure used to select and divide data in training, testing and evaluation steps. In the case of insufficient data, especially anomalies in a given dataset, training, evaluation and testing are performed using small number of data and even some classes may not be selected. To solve this problem, the following question should be answered: "Based on what criteria, the data are divided to testing, evaluation and training data?"

Random classification lowers the detection quality because different types of anomalies may be selected in the training or evaluation steps and this will decrease the quality of other stages and perhaps all samples of a type of anomaly may be selected in a stage. Our proposed method utilizes the same percentage to select different types of anomalies and data. For example, if 70% of the data is selected for training, a same ratio of a variety of data should be selected. For this purpose, DivideFcn function is defined in the program code to data. In order to select data at various stages of training, evaluation and testing on the basis of equal proportions this function is used. DivideFcn function is used to study the Iris dataset. It causes to increase the reliability of data selecting in each stage.

**Test NN Confusion Matrix**

| | | Target Class | | | |
|---|---|---|---|---|---|
| | 1 | 2 | 3 | 4 | Precision |
| **1** | 4 / 57.1% | 0 / 0.0% | 1 / 14.3% | 0 / 0.0% | 80.0% / 20.0% |
| **2** | 0 / 0.0% | 0 / 0.0% | 0 / 0.0% | 0 / 0.0% | NaN% / NaN% |
| **3** | 0 / 0.0% | 0 / 0.0% | 1 / 14.3% | 1 / 14.3% | 50.0% / 50.0% |
| **4** | 0 / 0.0% | 0 / 0.0% | 0 / 0.0% | 0 / 0.0% | NaN% / NaN% |
| Recall | 100% / 0.0% | NaN% / NaN% | 50.0% / 50.0% | 0.0% / 100% | 71.4% / 28.6% (Test Error) |

Figure 10 Neural network confusion matrix for the second dataset.

**Test GA Confusion Matrix**

| | | Target Class | | | |
|---|---|---|---|---|---|
| | 1 | 2 | 3 | 4 | Precision |
| **1** | 4 / 57.1% | 0 / 0.0% | 0 / 0.0% | 0 / 0.0% | 100% / 0.0% |
| **2** | 0 / 0.0% | 0 / 0.0% | 0 / 0.0% | 0 / 0.0% | NaN% / NaN% |
| **3** | 0 / 0.0% | 0 / 0.0% | 2 / 28.6% | 0 / 0.0% | 100% / 0.0% |
| **4** | 0 / 0.0% | 0 / 0.0% | 0 / 0.0% | 1 / 14.3% | 100% / 0.0% |
| Recall | 100% / 0.0% | NaN% / NaN% | 100% / 0.0% | 100% / 0.0% | 100% / 0.0% (Test Error) |

Figure 11. Genetic algorithm confusion matrix for the second dataset.

The third dataset used in this study is Iris dataset. Unlike previous datasets, Iris is a supervised dataset (Multi-Class). Therefore, the dataset is divided into 3 sub datasets according to the proposed framework section (Setosa, Versicolor and Virginica, because there are 3 classes). The two main features in this dataset include petal length and petal width because they show the highest distinction between the data globally. Below, the data in each sub dataset are labeled as shown in Tables 5 to 7.

**Table 5.** Labeling data in the third dataset, the first sub dataset.

| Cluster 3 | Cluster 2 | Cluster 1 | | |
|---|---|---|---|---|
| | | | 50 | #Point |
| #CNA 8 | #CNA 11 | #ND 15 | 3 | #Cluster |
| | | | 15 | #ND |
| | | | 19 | #CNA |
| | | | 9 | #CPA |
| | | | 7 | #PA |

**Table 6.** Labeling data in the third dataset, the second sub dataset.

| | | | 50 | #Point |
|---|---|---|---|---|
| Cluster 3 | Custer 2 | Cluster 1 | 3 | #Cluster |
| #CNA | #CNA | #ND | | |
| 5 | 8 | 20 | | |
| | | | 20 | #ND |
| | | | 13 | #CNA |
| | | | 6 | #CPA |
| | | | 11 | #PA |

**Table 7.** Labeling data in the third dataset, the third sub dataset.

| | | | | 50 | #Point |
|---|---|---|---|---|---|
| Cluster 4 | Cluster 3 | Custer 2 | Cluster 1 | 4 | #Cluster |
| #CNA | #CNA | #CNA | #ND | | |
| 4 | 8 | 6 | 17 | | |
| | | | | 17 | #ND |
| | | | | 18 | #CNA |
| | | | | 8 | #CPA |
| | | | | 7 | #PA |

Figure 12 shows distribution of 3 classes of Iris datasets in a 2-dimensional space.

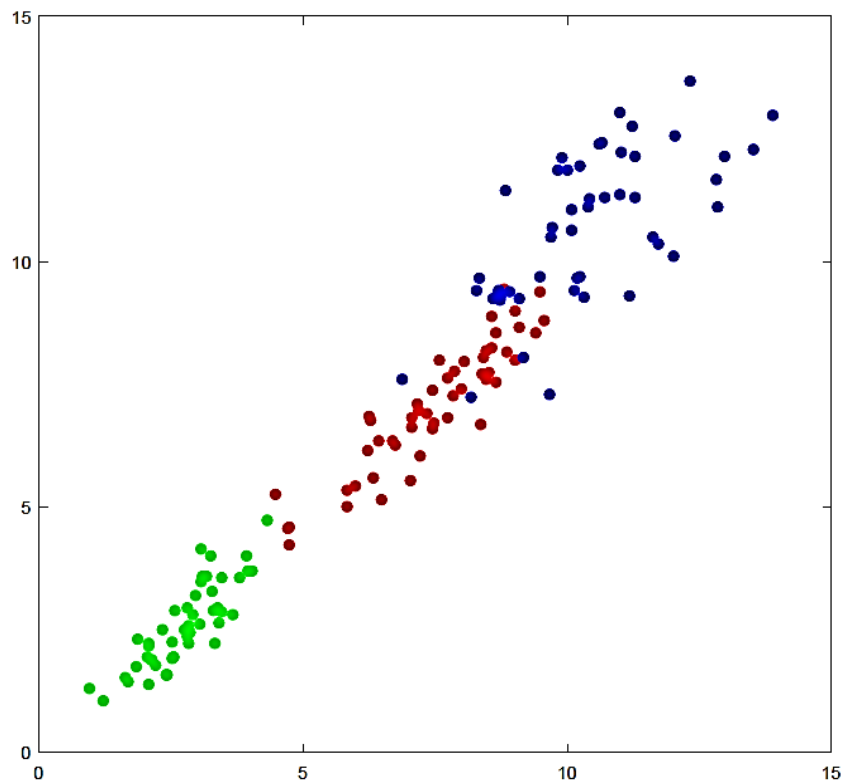

Figure 12. Iris dataset distribution in a 2-dimensional space, green: Setosa, red: Versicolor and blue: Virginica.

Figures13 and 14 show the confusion matrix for the Iris dataset. The matrix indicates the quality of proposed techniques in a supervised dataset. The thin boundary between anomalies in Class 2 and 3 in the conventional MLP-NN is unacceptable while the MLP-NN enhanced by the GA provides acceptable results.

Figure 13. Neural network confusion matrix for Iris dataset.

Figure 14. Genetic algorithm confusion matrix for Iris dataset.

Figure 15 shows the comparison of test error parameter based on enhanced and conventional MLP-NN.

At the end of this part the other standard evaluation metrics which called True Positive Rate (TPR/sensitivity) and False Positive Rate (FPR) are calculated based on Equations (14) to (16) and results are presented in Table 8. Additionally, Figures 16 to 18 show that the proposed method which enhanced by GA outperforms the conventional MLP-NN in term of ROC curve.

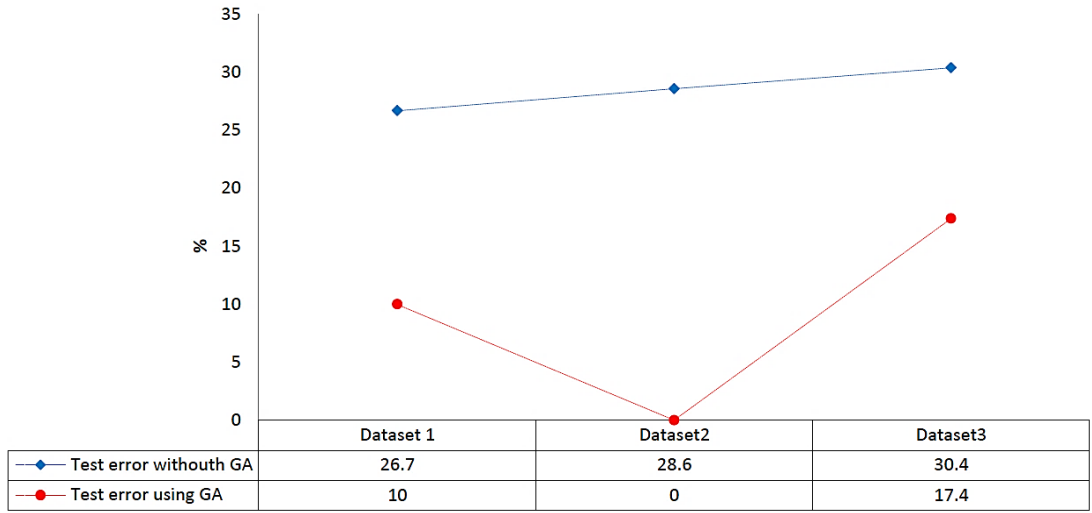

**Figure 15.** The comparison of test error parameter based on proposed framework.

$$TPR\ (Sensitivity) = \frac{TP}{TP+FN} \quad (14)$$

$$True\ Negative\ Rate(TNR\ or\ Specificity) = \frac{TN}{TN+FP} \quad (15)$$

$$FPR = 1 - TNR \quad (16)$$

**Table 8.** The comparison of TPR and FPR metrics in 3 Datasets basde on confusion matrixs.

| Classes | 1(ND) | | 2(CNA) | | 3(CPA) | | 4(PA) | |
|---|---|---|---|---|---|---|---|---|
| Datasets | TPR | FPR | TPR | FPR | TPR | FPR | TPR | FPR |
| D1-without GA | 1 | 0.166667 | 0 | 0 | 0.857143 | 0.2 | 0.666667 | 0.1 |
| D1-using GA | 1 | 0.0625 | 0.6 | 0 | 0.857143 | 0 | 1 | 0.086957 |
| D2-without GA | 1 | 0.5 | 0 | 0 | 1 | 0.2 | 0 | 0 |
| D2-using GA | 1 | 0 | 0 | 0 | 1 | 0 | 1 | 0 |
| D3-without GA | 1 | 0.363636 | 0.571429 | 0.2 | 0 | 0 | 1 | 0 |
| D3-using GA | 1 | 0.166667 | 0.714286 | 0.125 | 0.5 | 0 | 1 | 0 |

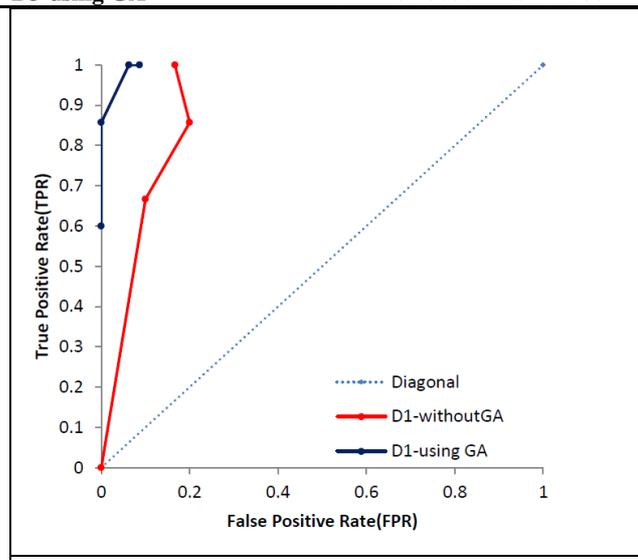

**Figure 16.** The ROC curve based on D1 dataset.

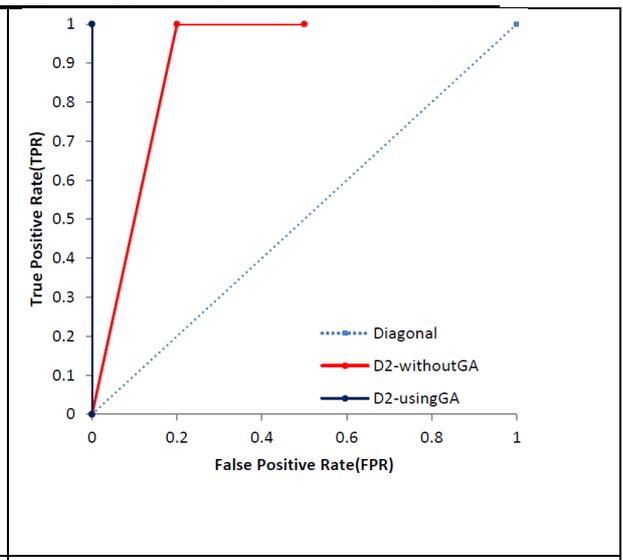

**Figure 17.** The ROC curve based on D2 dataset.

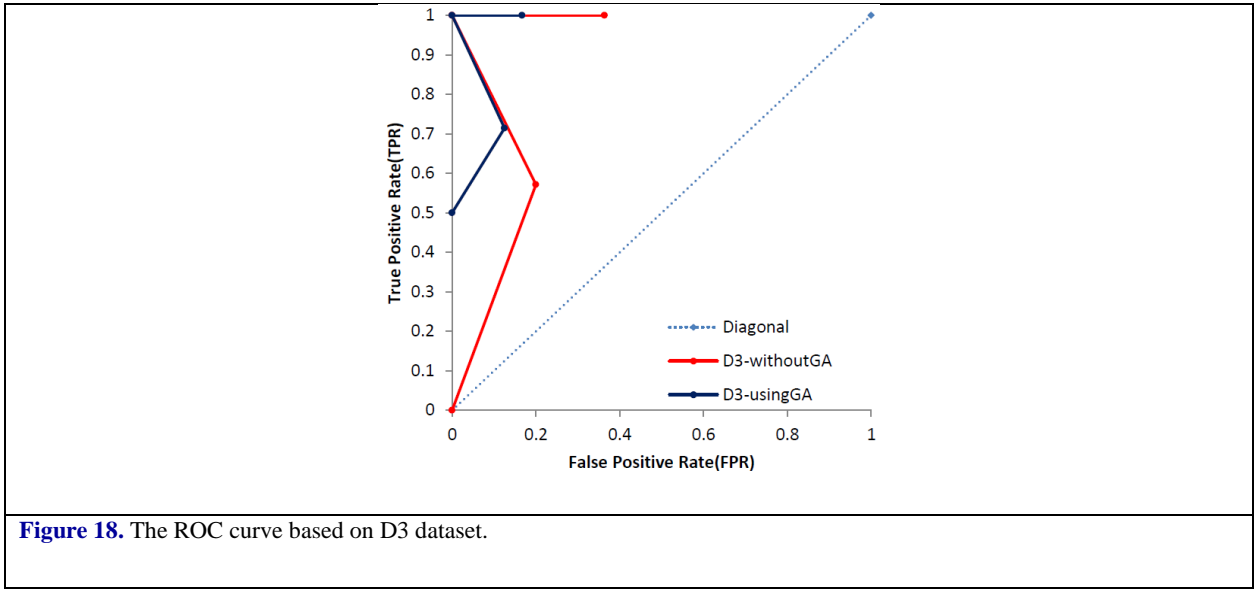

**Figure 18.** The ROC curve based on D3 dataset.

*Part B*

In this part, the proposed techniques have applied on 8 UCI datasets which presented the number of ND, CNA, CPA, PA and the thin boundary challenge between new anomalies including both CNA and CPA. Additionally, we have used a few benchmark datasets based on (Campos et al., 2016) repository which are more appropriate to consider ROC curve and results are presented in Figures 21 to 26.

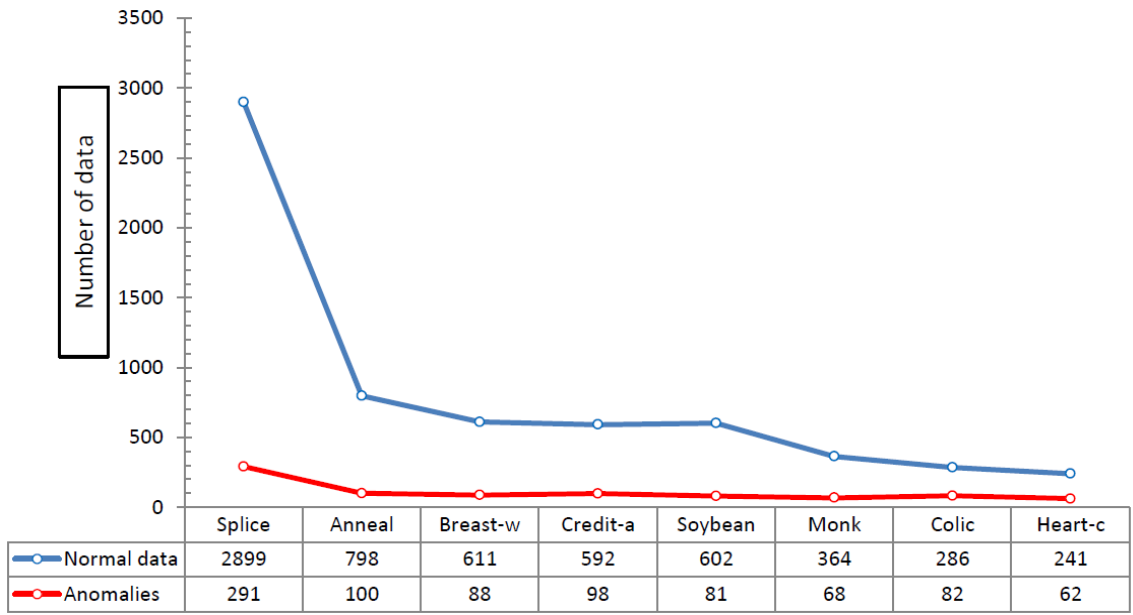

**Figure 19.** The thin boundary challenge between normal data and anomalies before new definition of anomalies overall 8 UCI datasets.

**Table 9.** Number of data overall 8 UCI datasets.

| Dataset name | #Cases | #Classes | #Attributes | #Cluster | #ND | #CNA | #CPA | #PA |
|---|---|---|---|---|---|---|---|---|
| Splice | 3190 | 3 | 62 | 12 | 2422 | 477 | 215 | 76 |
| Anneal | 898 | 6 | 39 | 9 | 452 | 346 | 68 | 32 |
| Breast-w | 699 | 2 | 9 | 7 | 412 | 199 | 63 | 25 |
| Credit-a | 690 | 2 | 15 | 7 | 418 | 174 | 71 | 27 |
| Soybean | 683 | 19 | 35 | 6 | 454 | 148 | 58 | 23 |
| Monk | 432 | 2 | 6 | 5 | 302 | 62 | 49 | 19 |
| Colic | 368 | 2 | 22 | 5 | 157 | 129 | 54 | 28 |
| Heart-c | 303 | 2 | 13 | 4 | 146 | 95 | 41 | 21 |

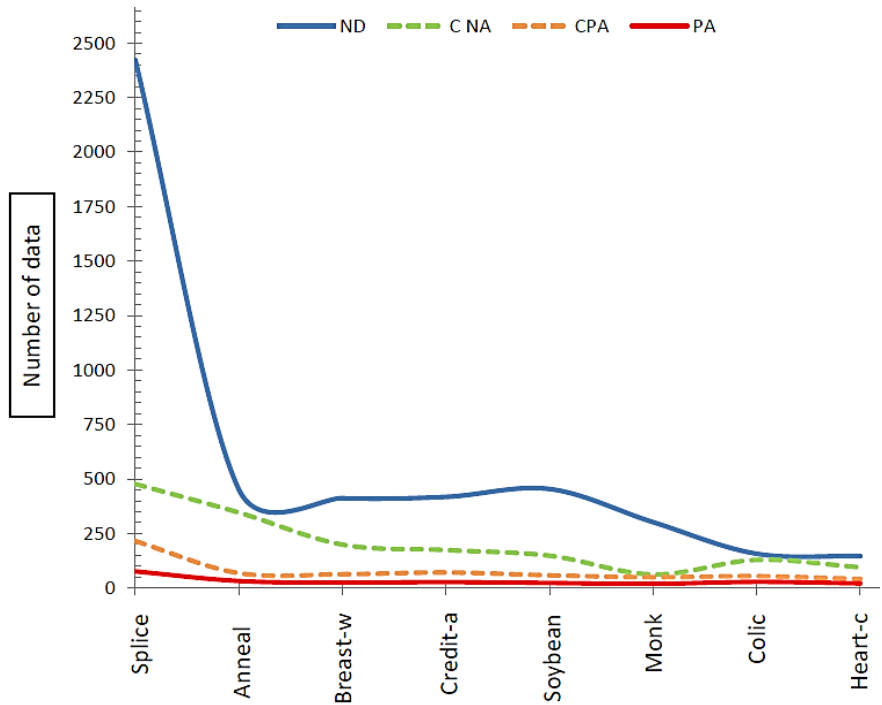

**Figure 20.** The thin boundary challenge between normal data and anomalies after new definition of anomalies overall 8 UCI datasets.

**Table 10.** Datasets which are available on (Campos et al., 2016) repository.

| Dataset name | Instances | Inliers | Outliers | Attributes |
|---|---|---|---|---|
| WPBC | 198 | 151 | 47 | 33 |
| Ionosphere | 351 | 225 | 126 | 32 |
| Waveform | 3443 | 3343 | 100 | 21 |
| Annthyroid | 7200 | 6666 | 534 | 21 |
| Pima | 768 | 500 | 268 | 8 |
| Parkinson | 195 | 18 | 147 | 22 |

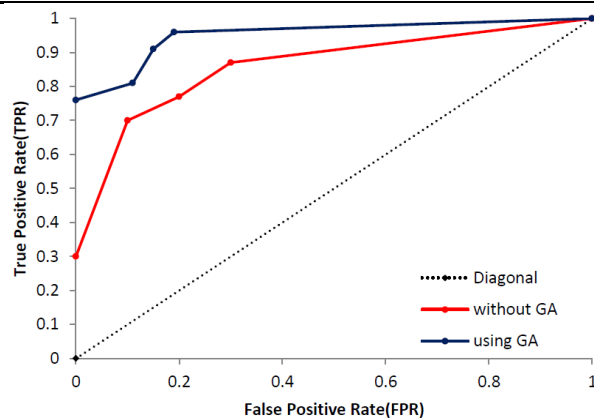

**Figure 21.** The ROC curve based on WPBC dataset.

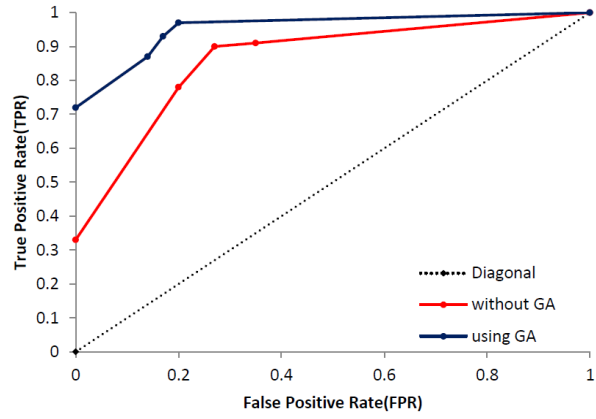

**Figure 23.** The ROC curve based on Waveform dataset.

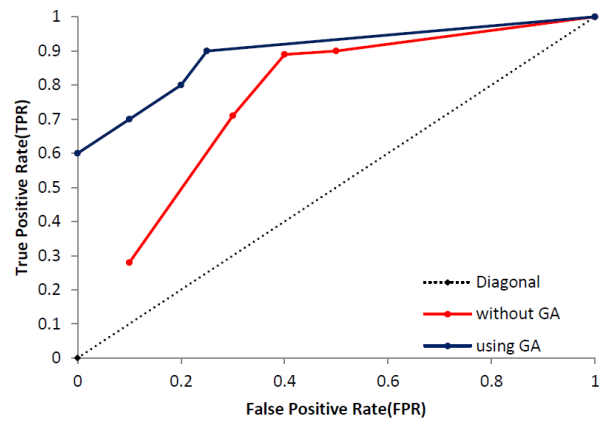

**Figure 25.** The ROC curve based on Pima dataset.

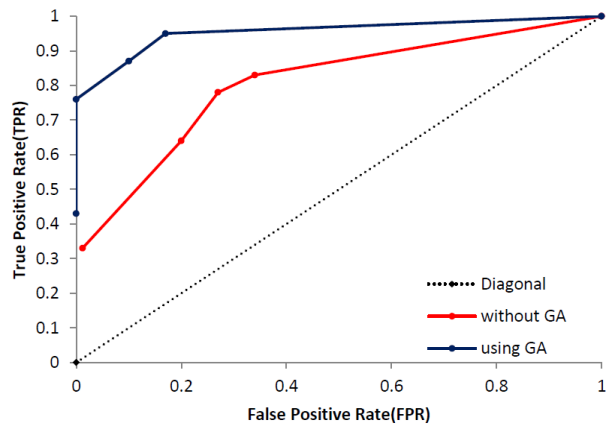

**Figure 22.** The ROC curve based on Ionosphere dataset.

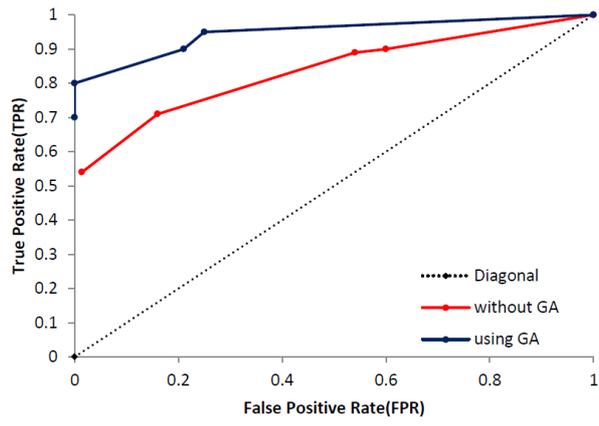

**Figure 24.** The ROC curve based on Annthyroid dataset.

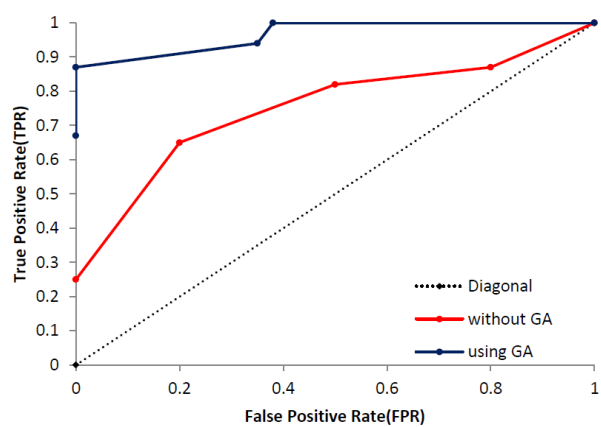

**Figure 26.** The ROC curve based on Parkinson dataset.

## Conclusion

The thin boundary between various types of anomalies was studied. For this purpose, a new framework was introduced to adapt the proposed approach to both supervised and unsupervised datasets. Then, the MLP-NN was enhanced using the GA to ensure a test error less than that calculated for the conventional MLP-NN. Moreover, new types of anomalies were investigated by applying the proposed method on benchmark datasets. The most important features of these methods include adaptability to both supervised and unsupervised datasets, improved detection of various types of anomalies, increased reliability and enhancement of MLP-NN by the GA.

In the future work, for comparing both the efficiency and effectiveness of the proposed algorithm large datasets in a special field (e.g., intrusion detection, credit card fraud detection) will be used. A suitable technique will be provided to inject outliers in a dataset with insufficient outliers. This result in high-quality division of data into training, evaluation and testing data and thus reduces the test error.